\documentclass[10pt,twocolumn,letterpaper]{article}

\usepackage{dv}
\usepackage{times}
\usepackage{epsfig}
\usepackage{graphicx}
\usepackage{amsmath}
\usepackage{amssymb}
\usepackage{subcaption}
\usepackage{amssymb}
\usepackage{booktabs}
\usepackage{algorithm}
\usepackage{algpseudocode}
\usepackage{lipsum,microtype}
\usepackage{graphbox}
\usepackage{soul}
\usepackage{wrapfig}
\usepackage{multirow}
\usepackage{arydshln}
\usepackage{xspace}
\usepackage[numbers]{natbib}
\usepackage[nolist]{acronym}
\usepackage{xcolor}
\usepackage{tabularx}
\usepackage{multirow}

\makeatother
\usepackage{tikz}
\def\myfigure#1#2{%
    \begin{figure}[tb]%
    \centering\includegraphics*[width = \linewidth]{#1}%
    \vspace{-.3cm}%
    \caption{#2}%
    \label{fig:#1}%
    \end{figure}%
}

\def\mycfigure#1#2{%
    \begin{figure*}[t]%
    \centering\includegraphics*[width = \linewidth]{#1}%
    \vspace{-.2cm}%
    \caption{#2}%
    \label{fig:#1}%
    \end{figure*}%
}

\soulregister\cref7
\soulregister\cite7
\soulregister\cite7
\soulregister\shortcite7
\soulregister\eg0
\soulregister\ie0
\soulregister\etal0

\newcommand{\mysection}[2]{\section{#1}\label{sec:#2}}
\newcommand{\mysubsection}[2]{\subsection{#1}\label{sec:#2}}

\newcommand{\refSec}[1]{Sec.~\ref{sec:#1}}
\newcommand{\refFig}[1]{Fig.~\ref{fig:#1}}
\newcommand{\refEq}[1]{Eq.~\ref{eq:#1}}
\newcommand{\refTab}[1]{Tab.~\ref{tab:#1}}
\newcommand{\refAlg}[1]{Alg.~\ref{alg:#1}}

\newcommand{\mymath}[2]{\newcommand{#1}{\TextOrMath{$#2$\xspace}{#2}}}

\usepackage[pagebackref,breaklinks,colorlinks]{hyperref}

\usepackage[capitalize]{cleveref}
\crefname{section}{Sec.}{Secs.}
\Crefname{section}{Section}{Sections}
\Crefname{table}{Table}{Tables}
\crefname{table}{Tab.}{Tabs.}

\DeclareGraphicsExtensions{.png,.jpg,.pdf,.ai,.psd}
\threedvfinalcopy %
\def\threedvPaperID{0300} %

\ifthreedvfinal\fi

\begin{acronym}
\acro{AABB}{Axis-Aligned Bounding Box}
\acro{CNN}{Convolutional Neural Network}
\acro{FID}{Frechet Inception Distance}
\acro{GAN}{Generative Adversarial Network}
\acro{NeRF}{Neural Radiance Field}
\acro{ReLU}{Rectified Linear Unit}
\acro{WGAN}{Wasserstein GAN}
\acro{SIFID}{Single Image \ac{FID}}
\end{acronym}

\definecolor{colorA}{HTML}{4285f4}
\definecolor{colorB}{HTML}{ea4335}
\definecolor{colorC}{HTML}{fbbc04}
\definecolor{colorD}{HTML}{34a853}
\definecolor{colorE}{HTML}{ff6d01}
\definecolor{colorF}{HTML}{46bdc6}
\definecolor{colorG}{HTML}{000000}
\definecolor{colorH}{HTML}{777777}

\usepackage{pifont}
\newcommand{\cmark}{\checkmark}%
\newcommand{\xmark}{\scalebox{0.85}{\ding{53}}}%

\newcommand{\scene}[1]{\textsc{#1}}
\newcommand{\supp}[1]{#1}
\newcommand{\method}[1]{\textcolor{color#1}{\texttt{#1}}}

\usepackage{adjustbox}
\newcolumntype{R}{%
    >{\adjustbox{angle=90}\bgroup}%
    l%
    <{\egroup}%
}

\newcommand{\myparagraph}[1]{\vspace{.15cm}\noindent\textbf{#1}\quad}

\newcommand{\name}{\textsc{3inGAN}\xspace}

\begin{document}

\title{\name: Learning a 3D Generative Model from  Images of a Self-similar Scene}

\author{Animesh Karnewar$^1$
\quad\quad
Oliver Wang$^2$
\quad \quad
Tobias Ritschel$^1$
\quad\quad
Niloy J. Mitra$^{1,2}$ \\
$^1$University College London
\quad\quad\quad
$^2$Adobe Research
}

\mymath{\inputImage}{I}
\mymath{\inputImages}{\mathcal I}
\mymath{\numberOfInputImages}{{n_\mathrm{2D}}}
\mymath{\featureGrid}{V}
\mymath{\featureGridResolution}{{n_\mathrm{V}}}
\mymath{\background}{B}
\mymath{\generativeModel}{G}
\mymath{\seed}{\mathbf z}
\mymath{\reconstructionSeed}{\seed^\star}
\mymath{\poses}{\mathcal C}
\mymath{\pose}{\mathsf C}
\mymath{\poseModel}{\mathcal D}\mymath{\rendering}{\mathcal R}
\mymath{\discriminator}{D}
\mymath{\generatorParameters}{\theta}
\mymath{\imageDiscriminatorParameters}{\phi}
\mymath{\gridDiscriminatorParameters}{\psi}
\mymath{\imageDiscriminator}{D_\mathrm{2D}}
\mymath{\gridDiscriminator}{D_\mathrm{3D}}
\mymath{\seedDimenstion}{n_\mathrm{z}}
\mymath{\patch}{P}
\mymath{\getPatch}{\mathcal P}
\mymath{\getImagePatch}{\getPatch_\mathrm{2D}}
\mymath{\getGridPatch}{\getPatch_\mathrm{3D}}
\mymath{\sample}{\mathtt{sample}}
\mymath{\weight}{\alpha}
\mymath{\imageCriticWeight}{\gamma_\mathrm{2D}}
\mymath{\gridCriticWeight}{\gamma_\mathrm{3D}}
\mymath{\imageReconstructionWeight}{\rho_\mathrm{2D}}
\mymath{\gridReconstructionWeight}{\rho_\mathrm{3D}}

\newcommand{\distribution}[2]{p_\mathrm{#1}^\mathrm{#2}}
\newcommand{\assign}{:=}
\newcommand{\learn}{\mathtt{up}}
\newcommand{\algospace}{\vspace{0.2cm}}
\newcommand\revision[1]{{#1}}

\twocolumn[{%
\renewcommand\twocolumn[1][]{#1}%
\maketitle
}]

\begin{abstract}
We introduce \name, an unconditional 3D generative model trained from 2D images of a single self-similar 3D scene.
Such a model can be used to produce 3D ``remixes'' of a given scene, by mapping spatial latent codes into a 3D volumetric representation, which can subsequently be rendered from arbitrary views using physically based volume rendering.
By construction, the generated scenes remain view-consistent across arbitrary camera configurations, without any flickering or spatio-temporal artifacts.  
During training, we employ a combination of 2D,  obtained through differentiable volume tracing, and 3D \ac{GAN} losses, across multiple scales, enforcing realism on both its 2D renderings and its 3D structure. 
We show results on semi-stochastic scenes of varying scale and complexity, obtained from real and synthetic sources. We demonstrate, for the first time, the feasibility of learning plausible view-consistent 3D scene variations from a single exemplar scene and provide 
qualitative and quantitative comparisons against two recent related methods. 
\revision{Code and data for the paper are available at} \url{https://geometry.cs.ucl.ac.uk/group_website/projects/2022/3inGAN/}.
\end{abstract}

\section{Introduction}

In the context of images, unconditional generative models, such as \acp{GAN}, learn to map latent spaces to diverse yet realistic high resolution images -- notable architectures include StyleGan~\cite{karras2019style} and BigGAN~\cite{brock2018large}. 
Furthermore, these models have been shown to contain high-level semantics in their latent space mappings, allowing powerful post-hoc image editing operations, such as changing the appearance and expression of a generated person~\cite{abdal2021styleflow, shen2020interfacegan, harkonen2020ganspace}. 
One key question therefore, is whether it is possible to learn similar generative models for \emph{3D} scenes.

\begin{figure}[t!]
    \centering
    \includegraphics[width=\columnwidth]{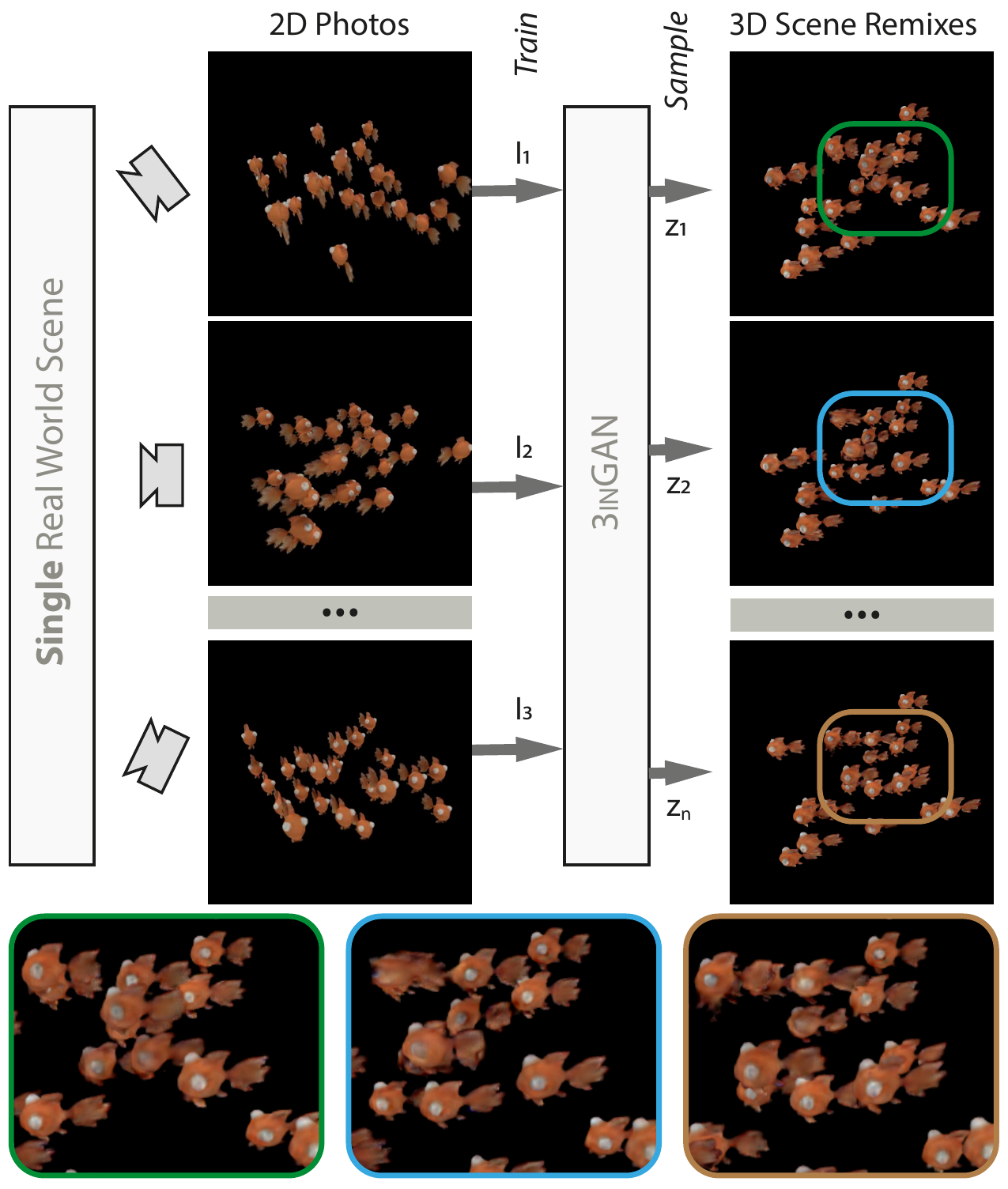}
    \caption{
    \textbf{Single scene 3D remixes. }
    We introduce \name that takes a set of 2D photos of a \textit{single} self-similar scene to produce a generative model of 3D scene remixes, each of which can be rendered from arbitrary camera configurations, without any flickering or  spatio-temporal artifacts. Bottom row insets show zooms from different generative samples, rendered from the same camera view, to highlight the quality and diversity of the results. }
    \label{fig:Teaser}
\end{figure}

There are, however, two key challenges. 
First, 3D generation suffers from \emph{data scarcity} as obtaining large and diverse datasets for 3D data, both geometry and appearance, is significantly more challenging than for 2D data, where one can simply scrape images from the internet.
Second, generative models struggle as the \emph{domain complexity} increases, as well as when datasets are not prealigned. 
This problem is even more severe in 3D, due to the added scene complexity, both in terms of scene structure and object appearance, and the fact that 3D models and scans often come with their own coordinate systems and/or scaling.

In this work, we propose \name, a solution to address both problems in a restrictive setting: an unconditional generative model for 3D scenes that works on a \emph{per-scene} basis for self-similar configurations. 
As our method does not require a large 3D dataset during training, it could be used across a wide range of real world domains.
By restricting the data domain to a \textit{single} 3D scene, we simplify the unconditional generation problem space to one with limited domain complexity, allowing us to learn a high-quality generator. 
We were inspired by similar approaches proposed for 2D images, e.g., SinGAN~\cite{shaham2019singan}. However, extending such approaches to 3D is nontrivial, as in 3D, one must be able to generate arbitrary views in a way that is multi-view consistent. 
We handle this by directly generating a 3D scene representation that is, by definition, multiview consistent.
In particular, we generate a regularly sampled nonlinearly-interpolated-voxel grid~\cite{Karnewar2022ReLUFields} due to its simplicity, local (feature) influence, rendering efficiency, and its amenability to the prevailing convolutional generator and discriminator architectures in 2D/3D.
In order to achieve realism, both in terms of 3D structure and 2D image appearance, we simultaneously use 3D feature patches and 2D image patches to obtain gradients from the 3D and 2D discriminators, respectively. 
We link the 3D and 2D domains by using a differentiable volume rendering module to interpret the feature grid as RGB images. 
Note that recent neural 3D generators (e.g., BlockGAN~\cite{nguyen2020blockgan},  GRAF~\cite{schwarz2020graf}, $\pi$-GAN~\cite{chan2021pi}) are trained on image/shape collections, and are not easily applicable in our setup (see Section~\ref{sec:Evaluation} for comparison). 
For example, \refFig{Teaser} shows a sampling of ``remixed" results generated from %
images of a school of fish.

In summary, ours is the first work to introduce an unconditional generative model from a \textit{single 3D scene}. 
In particular, we investigate scenes with some degree of stochastic structure, which are suitable to shuffling or ``remixing'' the scene content into a new 3D scene that makes sense. 
In addition, we make a further simplifying assumption in that we drop the view-specific effects  and reconstruct Lambertian scenes.
We evaluate our proposed method on a series of synthetic and real scenes and show that our approach outperforms baselines in terms of quality and diversity.

\section{Related Work}

\myparagraph{2D generative models.}
Generative modeling for image synthesis learns a distribution of colour values over the pixels of an image, and has seen tremendous progress recently, with GANs being the de-facto standard for synthesizing realistic looking images~\cite{karras2017progressive, karras2019style, karras2020analyzing, brock2018large, karnewar2020msg}. 
Recent works such as \cite{karras2020training} further improve result quality in data-sparse regimes, while \cite{karras2021alias} applies signal processing techniques to the generator architecture to correct for aliasing errors.
Other contenders for generative modelling include VAEs~\cite{vahdat2020nvae, alemi2018fixing, kingma2013auto, higgins2016beta}, flow-based models~\cite{kingma2018glow, dinh2016density, dinh2014nice}, noise-diffusion based models~\cite{song2021maximum, song2020improved, jolicoeur2020adversarial, aneja2020ncp}, and even hybrids of these methods \cite{pidhorskyi2020adversarial,aneja2020ncp,grover2018flow, pmlr-v48-larsen16}.
Key to training such methods is the availability of large scale image collections, and often times such works are used on specific target domains, such as portrait photos. 

More similar to ours, Single Image GANs (SinGANs) \cite{shaham2019singan,hinz2021improved} learn to generate the distribution of \emph{patches} of a single image, in a progressive coarse-to-fine manner so as to generate plausible variations from that one single image. 
Such approaches avoid the problem of needing a large dataset, while still enabling useful applications such as retargetting.
However, they are restricted to repeated, or stochastic-like patch-based variations. 
Ours shares similar advantages, as well as restrictions in terms of scene type as SinGAN. In Section~\ref{sec:Evaluation} we show that a na\"{i}ve extension to 3D volumes does not produce reasonable results.

\myparagraph{3D generative models.}
Due to the lack of large scale real-world datasets, much of the research in 3D generative modeling has stayed in the synthetic realm such as modeling only 3D shapes~\cite{wu2016learning,nash2020polygen,bouritsas2019neural, ben2018multi,cosmo2020limp,gadelha2020learning}, or materials~\cite{guo2020materialgan}.
Other methods use synthetic datasets for predicting scene structure and use differentiable rendering for %
end-to-end training~\cite{zhang2020image, kim2021drivegan, Ost_2021_CVPR, kar2019meta}. 
Recently, methods that directly model 3D scenes, either using explicit or implicit neural scene representations, have been gaining popularity~\cite{henzler2019escaping,nguyen2019hologan,chan2021pi, schwarz2020graf, niemeyer2021giraffe, unconstrained-scene-generation, zhou2021cips, gu2021stylenerf}. 
Another successful line of work~\cite{nguyen2019hologan, niemeyer2021giraffe} uses a neural renderer to render features from a volumetric grid, followed by per-image 2D CNNs used for upsampling, and as such are not multiview consistent. 
In contrast, \cite{chan2021pi, schwarz2020graf} and \cite{unconstrained-scene-generation} are view-consistent by design since they use an implicit neural representation and explicit-implicit neural representation respectively for the 3D structure, using a physically based rendering equation for obtaining the final 2D images.
Concurrent works such as \cite{zhou2021cips, gu2021stylenerf} produce impressive results performing 3D synthesis with multi-view consistency. 
Such approaches are designed to be trained only through 2D (image) supervision, and work best in cases where large datasets can be obtained for limited domains, such as faces or cars. 
Our method is also 3D view-consistent by design, 
and can be applied on a long-tail of real world scenes, provided they are self-similar. 

\mycfigure{overview}{
\textbf{\name setup. }
Overview of our approach with two parts: an initialization of a reference 3D feature grid (top) and a stage-wise learning of a generative model (bottom).
Input to the system is a set of 2D images seen on the top left.
From these, optimization using differentiable rendering for known views produces the reference feature grid, which is the input to the next step.
The rows below (``Level'') denote levels of training the generator, a 2D discriminator, and a 3D discriminator.
The 3D discriminator (right) gets random 3D patches from the reference or generated 3D grid, while the 2D discriminator (right) gets random 2D patches from reference or from generated renderings.
}

\myparagraph{Differentiable rendering.}
Differentiable rendering~\cite{tewari2020state} enables neural networks to be trained by losses on the resulting rendered images of a 3D representation, by allowing the gradient to be back-propagated through the network. 
This has shown to be a highly effective tool, especially for learning 3D representations that allow for novel view synthesis.
Multiple works~\cite{isola2017image, nguyen2020rgbd, aliev2020neural, wang2018high, mildenhall2019local, zhou2018stereo} 
have focused on designing neural-networks, either convolutional or otherwise, to go from spatial features to rendered pixels. 
Lately, methods that use volume tracing have been favored due to the advantage of flicker-free 2D rendering by design.
\acp{NeRF} \cite{mildenhall2020nerf} were the first to introduce this way of using differentiable rendering while using a neural 3D scene representation. Subsequently, many extensions of this have been proposed, to increase quality, robustness to scene type, or rendering speed~\cite{mildenhall2020nerf, zhang2020nerf++, barron2021mip, yu2021pixelnerf, chen2021mvsnerf, garbin2021fastnerf, yu2021plenoctrees, hedman2021baking, reiser2021kilonerf, fast-and-explicit-neural-view-synthesis}. 
Although using an MLP to learn a continuous scene representation has been shown to be able to lead to very high quality view synthesis results, such MLPs are not amenable in the generative contexts as ours, as explained next.

\mysection{Our Approach}{OurApproach}

\myparagraph{Motivation.}
As input, we require an exemplar self-similar scene, which is provided as a set of posed 2D images. 
We desire that our method produces a plausible 3D scene structure that matches the exemplar scene on a patch basis, and realistic 2D image renderings that are consistent across the space of all views of that scene.

Given this goal, we choose to directly generate a 3D representation, so that we are guaranteed view consistency when rendering 2D images from it. 
The first question is, \textit{what 3D representation should we use?} 
One choice would be coordinate-based MLPs, which have been shown to be compact scene representations able to generate very high quality novel views~\cite{mildenhall2020nerf}.
However, such a representation is ill-suited for a generative setup, as the costly evaluation makes rendering volumetric and image patches in the training loop infeasible, and the global and distributed nature of the MLP representation makes GAN training challenging in our patch-based setting. Another option is to operate directly on discretized RGBA volumes but, as demonstrated in Karnewar et al. \cite{Karnewar2022ReLUFields}, such an approach results in limited quality (blurry) results. Instead, we build our approach using their proposed grid-based ReLU Field representation. This representation is detailed in brief in section \ref{sec:relu_field}.

A na\"{i}ve extension of SinGAN~\cite{shaham2019singan} to 3D fails to produce good results for multiple reasons: (i)~only using a 3D discriminator doesn't have the notion of free-space in the volume and also tries to replicate the inside regions of the occupied space that may contain random values. When such a model is rendered, the results have minor shape distortions and chromatic noise; (ii)~only using 2D discriminators on the rendered images is challenging. Specifically,  when trained on too small patches the discriminator becomes too weak to inform the generator, and when trained on too large patches, the generator simply memorizes the initial reconstruction. 
Our solution involves two main ingredients: the use of ReLU-Fields representation \cite{Karnewar2022ReLUFields} instead of the standard RGBA volumes for inherently inducing the notion of free space in the 3D-grid; and a mixture of 2D and 3D discriminators, along with multi-scale training, that robustly produce consistent and high quality reconstructions.

\myparagraph{Method overview.}
Input to our method is a set of \numberOfInputImages 2D images $
\inputImages
:=
\{
\inputImage_1,
\ldots,
\inputImage_\numberOfInputImages 
\}
$ taken from one real world or synthetic self-similar scene.
\refFig{overview} presents a graphical overview and \refAlg{Main} a pseudo-code version.
Our method consists of two main steps.
First, we convert the 2D image set into a 3D feature grid \featureGrid (\refSec{Representation}).
Then, we train a generative model \generativeModel of 3D scenes from this 3D feature grid and its 2D rendered images (\refSec{Generation}).
This generative model $\generativeModel(\seed)$ then converts spatial random latent grids (\seed) into 3D feature grids containing remixes of the exemplar scene,  which can be consistently rendered from arbitrary views.

\begin{algorithm}[t!]
    \caption{Our \name training.
    Function $\mathtt{sample}(\ldots)$ samples all distributions provided as arguments; 
    $\learn(a,b)$ ADAM-updates the parameters $a$ by the gradient of expression $b$ with respect to $a$.}
    \label{alg:Main}
    \begin{algorithmic}[1]
    \Repeat    
        \Comment{3D representation building}
        \State $\{\inputImage,\pose\}\assign\sample(\inputImages,\poses)$
        \State $\featureGrid \assign \learn(\featureGrid, \|\rendering(\featureGrid, \pose)-\inputImage\|^2_2)$
    \Until{converged.}
    \algospace
    \Repeat     
    \Comment{Generator training}
        \State $\{\seed,\pose\}\assign\sample(\mathcal N, \poseModel)$
        \State $\patch\assign\getImagePatch(\rendering(\generativeModel^\generatorParameters(\seed), \pose))$
        \State $\imageDiscriminatorParameters\assign\learn(\imageDiscriminatorParameters,\imageDiscriminator^\imageDiscriminatorParameters(P=\mathtt{fake}))$
        \algospace
        \State $\pose\assign\sample(\poseModel)$
        \State $\patch\assign\getImagePatch(\rendering(\featureGrid, \pose))$
        \State $\imageDiscriminatorParameters\assign\learn(\imageDiscriminatorParameters,\imageDiscriminator^\imageDiscriminatorParameters(P=\mathtt{real}))$        
        \algospace
        \State $\seed\assign\sample(\mathcal N)$
        \State $\patch\assign\getGridPatch(\generativeModel^\generatorParameters(\seed))$
        \State $\gridDiscriminatorParameters\assign\learn(\gridDiscriminatorParameters,\gridDiscriminator^\gridDiscriminatorParameters(P=\mathtt{fake}))$
        \algospace
        \State $\patch\assign\getGridPatch(\featureGrid)$
        \State $\gridDiscriminatorParameters\assign\learn(\gridDiscriminatorParameters,\gridDiscriminator^\gridDiscriminatorParameters(P=\mathtt{real}))$
        \algospace
        \State $\{\seed,\pose\}\assign\sample(\mathcal N, \poseModel)$
        \State $\generatorParameters\assign\learn(\generatorParameters,\imageDiscriminator^\imageDiscriminatorParameters(\getImagePatch(\rendering(\generativeModel^\generatorParameters(\seed), \pose))=\mathtt{real}))$
        \State $\generatorParameters\assign\learn(\generatorParameters,\gridDiscriminator^\gridDiscriminatorParameters(\getGridPatch(\generativeModel^\generatorParameters(\seed))=\mathtt{real}))$
    \Until{converged.}
    \end{algorithmic}
\end{algorithm}

\mysubsection{Representation}{Representation}

\myparagraph{Foreground ReLU Field \cite{Karnewar2022ReLUFields}.}
\label{sec:relu_field}
Foreground is bound by a user-provided \ac{AABB} covered by a volumetric grid, \featureGrid, of fixed resolution $
\featureGridResolution_x
\times
\featureGridResolution_y
\times
\featureGridResolution_z
$ to contain feature values in the $[-1, 1]$ range. These values correspond to the raw-features that are stored on the voxel-grid. In order to obtain continuous density field, the trilinearly interpolated values of these raw features are passed through a single channel-wise \ac{ReLU} to convert them to $[0,1]$ range which can be physically interpreted as the density values. We do not model view-dependent appearance, \ie, we approximate the scene with Lambertian materials.

\myparagraph{Background.}
The background is assumed to be constant black for synthetic scenes.
For real scenes, we model the background of the scene using an implicit neural network \background, similar to NeRF++~\cite{zhang2020nerf++}, but without using the inverted-sphere parametrization of the scene.
Our goal is not to model the entire scene perfectly, but rather to provide appropriate inductive bias to the reconstruction pipeline to do the foreground-background separation correctly. 
This allows us to reconstruct real-world scenes without the requirement of additional segmentation masks.

\myparagraph{Optimization.}
Let the camera pose (extrinsic translation and rotation as well as intrinsics) for each input image be $
\poses :=
\{
\pose_1,
\ldots,
\pose_\numberOfInputImages
\}$ and assume they are known, \eg, by using structure-from-motion (we use ColMap~\cite{schonberger2016structure}).
Further, we denote the rendering operation to convert the feature grid \featureGrid and the camera pose \pose into an image as $\rendering(\featureGrid, \pose)$.
Specifically, we use emission-absorption raymarching~\cite{max1995optical,henzler2019escaping}. See supplementary for more details for the rendering.
We can then directly optimize the feature grid's photometric loss, given the pose and the 2D images as, 
\begin{equation}
\operatorname{arg\,min}_\featureGrid
\sum_{i=1}^\numberOfInputImages
\|
\inputImage_i-\rendering(\featureGrid, \pose_i)
\|^2_2.
\end{equation}

We minimize  with batched optimization over 2048 random rays out of all the rays for which we know the 2D input image pixel value. Input images are of size $512 \times 512$. 
Further, instead of directly optimizing for the full-resolution volume \featureGrid, training proceeds progressively in a coarse-to-fine manner.
Initially, the feature grid is optimized at a resolution where each dimension is smaller by factor 16.
After seeing 20k batches of input rays, the feature grid resolution is multiplied by two and the feature grid tri-linearly upsampled.

\mysubsection{Generation}{Generation}
Training the generative model makes use of the 3D feature grid \featureGrid trained in the previous section \ref{sec:Representation}, which we denote as the \emph{reference} grid herein.
We look into the generator details first, before explaining the losses used to train it: 2D and 3D discriminators, and a 2D and 3D reconstruction loss.

\myparagraph{Generator.}
Recall, that the model \generativeModel maps random latent codes \seed to a 3D feature grid $\generativeModel(\seed)$ at the coarsest stage. While adds fine residual details to previous stage's outputs at the rest of the stages similar to SinGAN \cite{shaham2019singan}.
The generator is a 3D CNN that stagewise decodes a spatial grid of noise vectors \seed of size \seedDimenstion (we use $seedDimension=4$) into the grid of the desired resolution. 

\myparagraph{Training.}
We train the architecture progressively: the generator first produces grids of reduced resolution.
Only once this has converged, layers are added and the model is trained to produce the higher resolution. Note that we freeze the previously trained layers in order to avoid the GAN training from diverging.
We employ an additional reconstruction loss that enforces one single fixed seed \reconstructionSeed to map to the reference grid. We supervise this fixed seed loss via an MSE over the 3D grids and with 2D rendered patches.

\myparagraph{2D discriminator.}
A 2D loss discriminates 2D patches of renderings of the generated feature grid to 2D patches rendered from the reference grid \featureGrid.
To render the 3D grids we need to model another distribution of poses, denoted by \poseModel, that uniformly samples camera locations to point at the center of the hemisphere and where focal length is varied stagewise linearly, where the value at the final stage corresponds to the actual camera intrinsics.

Further, let $\getImagePatch()$ be an operator to extract a random patch from a 2D image, with discriminators, 
\begin{equation}
\distribution{F}{2D}=
\getImagePatch(\rendering(\generativeModel(\seed), \poseModel))
\text{\; and \;}
\distribution{R}{2D}=
\getImagePatch(\rendering(\featureGrid, \poseModel)).
\end{equation}
Note, that we did \textit{not} define $
\distribution{R}{2D}=
\getImagePatch(\inputImages)
$, as this would limit ourselves to use real samples only from the limited set of known 2D image patches.
``Trusting'' our reference 3D feature grid has been extracted properly, we can instead sample it from arbitrary views and get a much richer set.
 
\myparagraph{3D discriminator.}
The 3D discriminator compares 3D patches from the generated feature grid to 3D patches of the reference feature grid.
Let $\getGridPatch(\featureGrid)$ be an operator to extract a random patch from a 3D feature grid \featureGrid.
The distributions to discriminate are,  \begin{equation}
\distribution{F}{3D}=
\getGridPatch(\generativeModel(\seed))
\text{\;\; and \;\;}
\distribution{R}{3D}=
\getGridPatch(\featureGrid).
\end{equation}
Finally, we use Wasserstein GAN~\cite{arjovsky2017wasserstein} to both distributions as well as the reconstruction losses (in both 2D and 3D),
\begin{equation}
\begin{aligned}
    \label{eq:Loss}
    \mathcal{L}=    
    \imageCriticWeight\cdot
    &\mathtt{wgan}(\distribution{R}{3D}, \distribution{F}{3D})+
    \gridCriticWeight\cdot
    \mathtt{wgan}(\distribution{R}{2D}, \distribution{F}{2D})+\\
    \imageReconstructionWeight\cdot
    &\mathbb E_{\pose\in\poseModel}
    [\|
    \rendering(\generativeModel(\reconstructionSeed), \pose)-
    \rendering(V, \pose)
    \|_2^2]+\\
    \gridReconstructionWeight\cdot
    &\|\generativeModel(\reconstructionSeed)-V\|_2^2
    ,
\end{aligned}    
\end{equation}
weighted by two pairs of two factors \imageCriticWeight, \gridCriticWeight and \imageReconstructionWeight, \gridReconstructionWeight.
In practice, we use $\imageCriticWeight=\gridCriticWeight=1.0$ and as well as 
$\imageReconstructionWeight=\gridReconstructionWeight=10$.

\mysection{Evaluation}{Evaluation}
We perform quantitative and qualitative evaluations of our approach on a number of different scenes, on which we compare to baseline methods, and evaluate design choices via an ablation study. 

\colorlet{colorOurs}{colorA}
\colorlet{colorPiGAN}{colorB}
\colorlet{colorGraf}{colorC}
\colorlet{colorOursSinGAN3D}{colorD}
\colorlet{colorOursPlatoGAN}{colorE}

\myparagraph{Comparisons and ablations.}
We compare to four methods (\refTab{Baselines}). 
Besides our full method \name (\method{Ours}), we study two prior approaches and two ablations.

As there is no existing method for 3D single scene remixing, we instead compare to two recent methods that were designed to learn a 3D generative model for \emph{classes} of objects, trained on a dataset of images/renderings where each image corresponds to a different instance and view, \method{PiGAN}~\cite{chan2021pi} and \method{Graf}~\cite{schwarz2020graf}. 
In these baselines, we test how well such methods work when given instead, many rendered views of a single scene.
In both cases, we use the code provided by the authors and their recommended parameter settings.
We mark whether the original approach was designed for a single scene or not in the last column in \refTab{Baselines}.

In the first ablation, our 2D only ablation, we evaluate the importance of the 3D discriminator and 3D reconstruction seed losses 
by running a version of our method with those losses removed, i.e. ($\gridReconstructionWeight=\gridCriticWeight=0$ in \refEq{Loss}).
We refer to this method as \method{OursPlatoGAN}, as its GAN loss setting, which is only on the rendered 2D images, is similar to the setup used in Henzler et al.~\cite{henzler2019escaping}, while being applied on only a single scene.

In the second ablation, our 3D only ablation, we evaluate the importance of the 2D discriminator and 2D reconstruction seed loss, which we refer to as \method{OursSinGAN3D}.
This approach is a na\"{i}ve extension of SinGAN~\cite{shaham2019singan} to 3D.
In other words, it is our approach without any differentiable rendering, i.e., with the 2D discriminator and 2D reconstruction seed losses removed ($\imageCriticWeight=\imageReconstructionWeight=0$).
\supp{Please see the supplementary material for further ablations of the reconstruction seed losses.}

\mycfigure{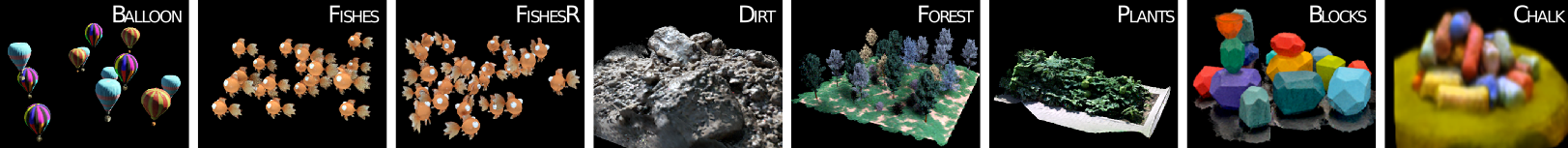}{
\textbf{Datasets. }
Example renderings of the scenes from our synthetic and real world datasets (\scene{Blocks}, \scene{Chalk}).}

\begin{table}[t!]
    \setlength{\tabcolsep}{8pt}
    \definecolor{xmarkcolor}{gray}{0.8} 
    \renewcommand{\xmark}{
    \color{xmarkcolor}\scalebox{0.85}{\ding{53}}
    }
    \centering
    \caption{
    \textbf{Comparisons and ablations.}
    We enumerate the different methods based on how they make use of 2D versus 3D information, and if they operate on a single scene or multiple scenes. 
    }
    \label{tab:Baselines}
    \resizebox{\linewidth}{!}{
    \begin{tabular}{r ccccc}
        \toprule
        \small 
        &
        \multicolumn2c{Disc.}&
        \multicolumn2c{Recon.}&
        \multicolumn1c{\multirow{2}{1cm}{Single scene}}
        \\
        \cmidrule(lr){2-3}
        \cmidrule(lr){4-5}
        & 2D & 3D & 2D & 3D & \\
        \midrule
        \method{PiGAN}~\cite{chan2021pi}&
        \cmark&
        \xmark&
        \xmark&
        \xmark&
        \xmark\\
        \method{Graf}~\cite{schwarz2020graf}&
        \cmark&
        \xmark&
        \xmark&
        \xmark&
        \xmark\\
        \method{OursPlatoGAN}&
        \cmark&
        \xmark&
        \cmark&
        \xmark&
        \cmark\\  
        \method{OursSinGAN3D}&
        \xmark&
        \cmark&
        \xmark&
        \cmark&
        \cmark\\
        \method{Ours}&
        \cmark&
        \cmark&
        \cmark&
        \cmark&
        \cmark\\
        \bottomrule
    \end{tabular}
    }
    \label{tab:ablations}
    \vspace{-.1cm}
\end{table}

\myparagraph{Scenes.}
We consider a mix of synthetic and real scenes, with various levels of stochasticity (a requirement for patch-based remixing of scenes).
These scenes include a synthetic scene rendered from Blender showing fishes with the same orientation
(\scene{Fish}) as well as with random orientations (\scene{FishRot}), a scene composed of 3D balloons (\scene{Balloons}), inspired by \cite{shaham2019singan}.
We also use four semi-synthetic scenes that were real scenes reconstructed from images using photogrammetry and then cleaned by an artist and sold on SketchFab: a pile of dirt (\scene{DirtPile}), a log pile (\scene{Logs}), and a bush (\scene{Plants}). We also make a fully synthetic (\scene{Forest}) scene which has a ground plane so as to resemble most real-world settings.
Finally, we include two real scenes with background for which we have no ground truth 3D available, which both show a random arrangement of geometric toys (\scene{Blocks}) or pieces of colored chalks (\scene{Chalk}).
Note that in all the cases, regardless of the source, our method only accesses 2D renderings/images of the scene, not the 3D scene. 

\myparagraph{Evaluation metric.}
We evaluate our method along two axes -- \emph{visual quality}, and \emph{scene diversity}.
With traditional 2D GANs, these are most commonly evaluated using \ac{FID}.
However, this metric is typically used over two datasets of images, whereas our situation is slightly different; we have only one ground truth scene, and a diverse distribution of generated scenes. 
We extend \ac{SIFID}~\cite{shaham2019singan}   to 3D, and also explicitly separate quality and diversity to separately compare along each axis.

\textit{Visual quality} is measured as the expectation of \ac{SIFID} scores between the distribution of exemplar 2D \emph{images} and the distribution of rendered generated 2D images for a fixed camera over multiple seeds. We compute this expectation by taking a mean over a number of camera-views.  
This is similar in sprit to \ac{SIFID}, except we compute it over images rendered from different views of the 3D scenes. Lower distance reflects better quality.
\begin{figure}[h!] 
    \centering
    \includegraphics[width=\columnwidth]{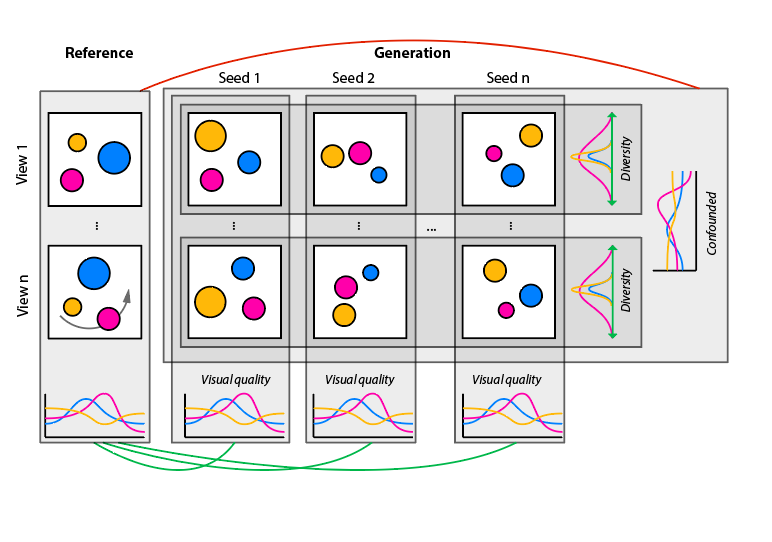}
    \caption{
    \textbf{Single Image FID.} 
    We extend \ac{FID} scores to our single scene use case.
The distribution of feature responses is computed for different camera views (rows) and generations (columns)
The reference (left) leads to a certain distribution of features.
Rather than matching the reference distribution across all views and all seeds (red lines), we compare it to the distribution of a single fixed seed across all views (green lines) to measure \textit{visual quality}.
Then, we compare the variance of the distribution of features across all seeds under a fixed view as a measure of \textit{scene diversity}.
}
    \label{fig:Metric}
\end{figure} 

Unfortunately, in the single scene case, we \textit{cannot} meaningfully compute \ac{FID} scores between the exemplar patch distribution and the distribution of all generated patches across seeds as it would make diversity appear as a distribution error. 
This is different from a typical GAN setup where we are given real images as samplings of desirable distribution of both quality and diversity. 
Instead, we measure \textit{scene diversity}  as the variance of a fixed patch from a fixed view over random seeds, which is a technique used to study texture synthesis diversity~\cite{henzler2020neuraltexture}.
Larger diversity is better.

\begin{table*}[t]
    \newcommand{\best}[1]{\textbf{#1}}
    \newcommand{\secondbest}[1]{\underline{#1}}
    
    \setlength{\tabcolsep}{7.0pt}
    \centering
    \caption{
    \textbf{Quality versus diversity.} 
    A good generative model should have a good mix of quality and diversity -- excellent quality with no diversity or vice versa are both undesirable. 
    Visual Quality and Scene Diversity for different methods (columns) and different data sets (rows). To simplify comparison, we normalize the numbers so that ours is always $1$. The best for each metric on each dataset is \textbf{bolded} and second best is \underline{underlined}. Please refer to the supplementary for unscaled numbers. }
    \label{tab:Results}
    \begin{tabular}{r rr rr rr rr rr}
    \toprule
    &
    \multicolumn2c{\method{PiGAN}~\cite{chan2021pi}}&    
    \multicolumn2c{\method{Graf}~\cite{schwarz2020graf}}&
    \multicolumn2c{\method{OursPlatoGAN}}&
    \multicolumn2c{\method{OursSinGAN3D}}&
    \multicolumn2c{\method{Ours}}
    \\
    \cmidrule(lr){2-3}
    \cmidrule(lr){4-5}
    \cmidrule(lr){6-7}
    \cmidrule(lr){8-9}
    \cmidrule(lr){10-11}
    &
    \multicolumn1c{\footnotesize Qual. $\downarrow$}&
    \multicolumn1c{\footnotesize Div. $\uparrow$}&
    \multicolumn1c{\footnotesize Qual. $\downarrow$}&
    \multicolumn1c{\footnotesize Div. $\uparrow$}&
    \multicolumn1c{\footnotesize Qual. $\downarrow$}&
    \multicolumn1c{\footnotesize Div. $\uparrow$}&
    \multicolumn1c{\footnotesize Qual. $\downarrow$}&
    \multicolumn1c{\footnotesize Div. $\uparrow$}&
    \multicolumn1c{\footnotesize Qual. $\downarrow$}&
    \multicolumn1c{\footnotesize Div. $\uparrow$}\\    
    \midrule
\scene{Fish} & 15.74 & 0.261 & 264.27 & 0.359 & 465.47 & \best{2.282} & \secondbest{8.98} & 0.332 & \best{1.00} & \secondbest{1.000} \\
\scene{FishR} & 2.47 & 0.440 & 3.07 & \secondbest{1.018} & 46.02 & \best{2.499} & \secondbest{2.70} & 0.170 & \best{1.00} & 1.000 \\
\scene{Balloons} & \secondbest{1.44} & 0.024 & 3.25 & 0.477 & 1.79 & 0.482 & 9.57 & \secondbest{0.083} & \best{1.00} & \best{1.000} \\
\scene{Dirt} & \secondbest{0.96} & 0.264 & 1.69 & \secondbest{0.440} & \best{0.66} & 0.131 & 6.06 & 0.164 & 1.00 & \best{1.000} \\
\scene{Forest} & 1.33 & 0.473 & 1.69 & 0.801 & \best{0.70} & \secondbest{0.883} & 2.14 & 0.439 & \secondbest{1.00} & \best{1.000} \\
\scene{Plants} & \secondbest{5.99} & 0.261 & 6.56 & \secondbest{0.648} & 12.81 & 0.144 & 9.15 & 0.326 & \best{1.00} & \best{1.000} \\
\scene{Blocks} & \secondbest{0.91} & 0.224 & \best{0.73} & \secondbest{0.342} & 1.83 & 0.334 & 3.32 & 0.329 & 1.00 & \best{1.000} \\
\scene{Chalk} & \secondbest{0.02} & 0.061 & \best{0.01} & \secondbest{0.320} & 0.54 & 0.088 & 0.85 & 0.035 & 1.00 & \best{1.000} \\
    \bottomrule
\end{tabular}
\vspace{-0.2cm}
\end{table*}

\mycfigure{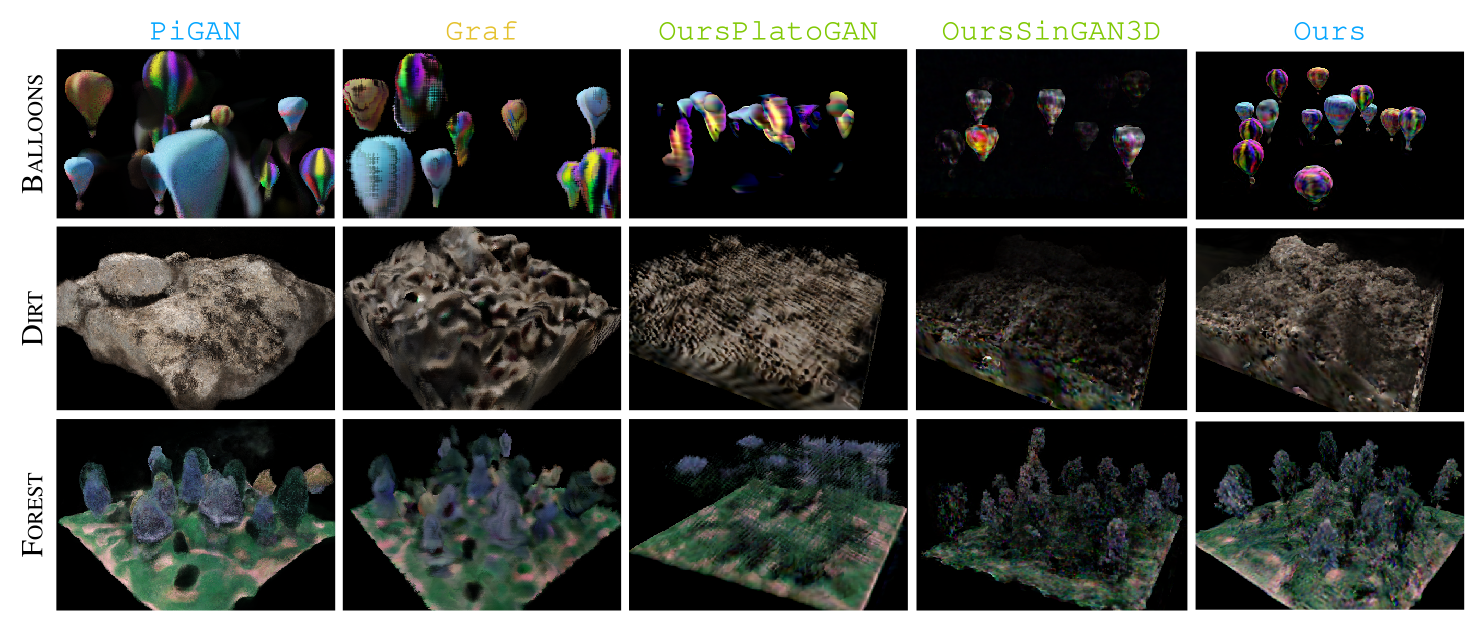}{
\textbf{Qualitative comparison.} 
Comparison of visual quality for different methods (columns) for different scenes (rows).}

\mycfigure{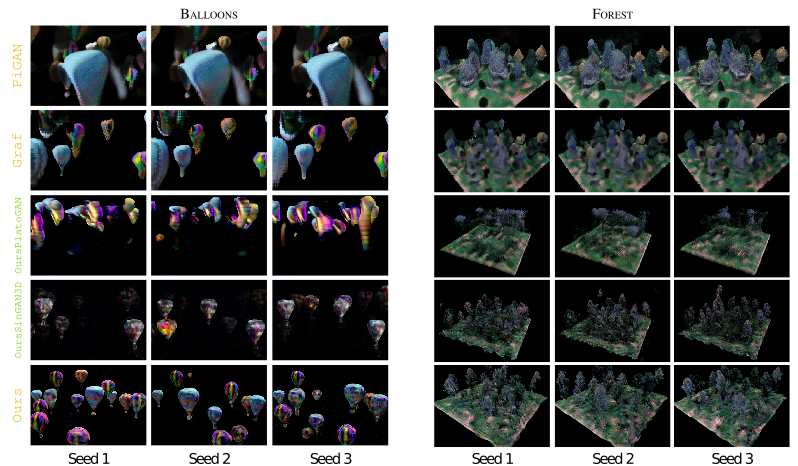}{
\textbf{Diversity across different generative samples.}   
Diversity under changing seeds (columns) of different methods (rows) for different scenes (left and right blocks).
See also Figure~\ref{fig:Teaser}.
}

\subsection{Qualitative Evaluation}

In Fig.~\ref{fig:quality_results} and Fig~\ref{fig:diversity_results}, we present qualitative results of \name, prior methods, and ablations.
\supp{Please see the supplementary material for video results that can better visually demonstrate the quality  and diversity of reconstructed scenes.}
We can see that, by construction, our method produces variations of 3D scenes that are view consistent.

\myparagraph{Visual quality.} 
While some artifacts remain, we observe that this task is challenging for all existing approaches, and when compared to prior work and simple baselines, our method achieves higher visual quality.
As expected, scenes with higher stochasticity result in better visual quality and diversity.
For example, the balloons are mostly intact but shifted to different locations, and the dirt pile consists largely of reasonable structures. 

\myparagraph{Scene diversity.}
Compared to the competing baselines,  \method{PiGAN} and \method{Graf}, we observe that
\name obtains significantly more scene diversity.
This is not surprising, as prior methods have been designed to work on multi-scene image collections, and hence experience mode collapse when given views of only a single scene.
This motivates the development of a patch-based generative model for 3D scenes.
We see that \name learns to keep the identity of objects (e.g., balloons, fish, chalk) and create plausible 3D variations (e.g., balloons floating in air, or blocks meaningfully stacked). \revision{Please refer to the supplementary for more results.} 

\myfigure{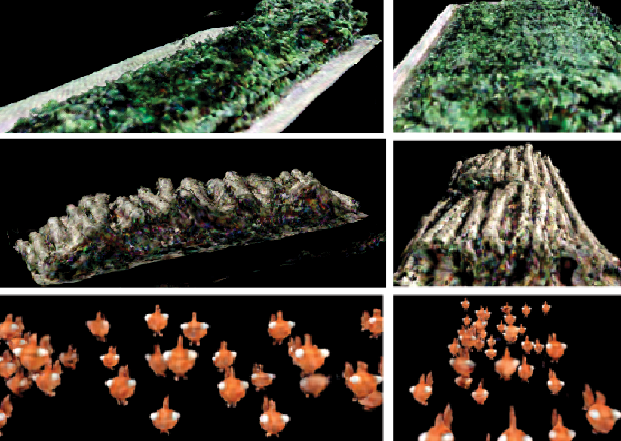}{
\textbf{Scene retargetting.} 
Retargeting the \scene{Plants}, the \textsc{Logs}, and \scene{Fish}  scenes to novel aspect ratios.
Since ours is CNN-based, it is easy to retarget scenes to different sizes. 
}
\myparagraph{Scene retargeting.}
In \refFig{retargeting} we show results of changing the aspect ratio of generated scenes.
As our generator is fully convolutional, this can be done simply by changing the shape of the input noise. 
We can see that the scene structure remains plausible, with the semi-stochastic content repeated to fill in the space.

\subsection{Quantitative Evaluation}

In \refTab{Results}, we quantitatively compare the different variations using our proposed metrics.
Mirroring the qualitative observations, we can see that \name performs better than prior work and ablations in terms of visual quality.
The numbers show the importance of having both the 2D and 3D discriminators as they help to improve the object appearance and geometric structure, respectively, of the generations. 
Our main advantage is the noticeable boost in diversity of the generated results, as also measured in the 2--5$\times$ boost in scene diversity score over the other methods. 
Note that as indicated by the relative \textit{Visual Quality} scores, our results are not yet close to being photorealistic (indistinguishable from the reference distribution) but nonetheless we achieve a healthy gain in over quality/diversity over competing alternatives.

\section{Discussion}

\myparagraph{Limitations and future work.}
Our approach has a number of limitations. 
For one, we note that the proposed method of generating a distribution (i.e., a generative model) from a single scene only makes sense in case where scenes contain stochastic structures that can be shuffled around.
We do not expect our method to work on highly structured content, e.g., people.
Furthermore, we observe that results can still exhibit artifacts, such as high frequency noise, blobby output, and floaters.
We believe that this is because the scene distribution is being estimated from very limited data (i.e., a single scene), and is analogous to the ``splotchyness'' commonly seen in 2D GANs when trained in limited data settings or over complex distributions.
In this case, it should be possible to improve quality by using data augmentation as recommended in~\cite{karras2020training}, or by improving the training regime, as in~\cite{hinz2021improved}, however in the latter case it is important to balance diversity to avoid mode collapse.
Finally, in our implementation, we assume scenes to be Lambertian, i.e., without view-dependent effects.
In the future, view-dependent specular effects could be modelled using SH components. %

\myparagraph{Conclusion.}
We have presented \name, a method for training a generative model of remixes of a single 3D scene.
Similar to SinGAN, our approach works on ``long-tail" data, as it does not require aligned 3D datasets for training. 
Instead, we require only a set of posed images of a single self-similar scene as input. 
We believe that this method has a number of downstream applications, such as retargeting or harmonization~\cite{shaham2019singan}, and furthermore, studies in generative models for 3D scenes may, one day, lead to realistic view-consistent scene generation on par with images.
This would be practical not only for 3D content generation, but as a generative prior to be used in many 3D reconstruction and editing tasks.

\paragraph{Acknowledgements.}
\revision{
The work was partially supported by the Marie Skłodowska-Curie grant agreement No. 956585, gifts from Adobe, and the UCL AI Centre.}

\setlength{\bibsep}{0.7pt}
{\small
\bibliographystyle{ieee_fullname}
\bibliography{main_dv}

\begin{thebibliography}{10}\itemsep=-1pt

\bibitem{abdal2021styleflow}
Rameen Abdal, Peihao Zhu, Niloy~J Mitra, and Peter Wonka.
\newblock Styleflow: Attribute-conditioned exploration of stylegan-generated
  images using conditional continuous normalizing flows.
\newblock {\em ACM Trans. Graph.}, 40(3):1--21, 2021.

\bibitem{alemi2018fixing}
Alexander Alemi, Ben Poole, Ian Fischer, Joshua Dillon, Rif~A Saurous, and
  Kevin Murphy.
\newblock Fixing a broken elbo.
\newblock In {\em International Conference on Machine Learning}, pages
  159--168. PMLR, 2018.

\bibitem{aliev2020neural}
Kara-Ali Aliev, Artem Sevastopolsky, Maria Kolos, Dmitry Ulyanov, and Victor
  Lempitsky.
\newblock Neural point-based graphics.
\newblock In {\em ECCV}, pages 696--712. Springer, 2020.

\bibitem{aneja2020ncp}
Jyoti Aneja, Alexander Schwing, Jan Kautz, and Arash Vahdat.
\newblock Ncp-vae: Variational autoencoders with noise contrastive priors.
\newblock {\em arXiv preprint arXiv:2010.02917}, 2020.

\bibitem{arjovsky2017wasserstein}
Martin Arjovsky, Soumith Chintala, and Léon Bottou.
\newblock Wasserstein gan, 2017.

\bibitem{barron2021mip}
Jonathan~T Barron, Ben Mildenhall, Matthew Tancik, Peter Hedman, Ricardo
  Martin-Brualla, and Pratul~P Srinivasan.
\newblock Mip-nerf: A multiscale representation for anti-aliasing neural
  radiance fields.
\newblock {\em arXiv preprint arXiv:2103.13415}, 2021.

\bibitem{ben2018multi}
Heli Ben-Hamu, Haggai Maron, Itay Kezurer, Gal Avineri, and Yaron Lipman.
\newblock Multi-chart generative surface modeling.
\newblock {\em ACM Trans. Graph.}, 37(6):1--15, 2018.

\bibitem{bouritsas2019neural}
Giorgos Bouritsas, Sergiy Bokhnyak, Stylianos Ploumpis, Michael Bronstein, and
  Stefanos Zafeiriou.
\newblock Neural 3d morphable models: Spiral convolutional networks for 3d
  shape representation learning and generation.
\newblock In {\em ICCV}, pages 7213--7222, 2019.

\bibitem{brock2018large}
Andrew Brock, Jeff Donahue, and Karen Simonyan.
\newblock Large scale gan training for high fidelity natural image synthesis.
\newblock {\em arXiv preprint arXiv:1809.11096}, 2018.

\bibitem{chan2021pi}
Eric~R Chan, Marco Monteiro, Petr Kellnhofer, Jiajun Wu, and Gordon Wetzstein.
\newblock pi-gan: Periodic implicit generative adversarial networks for
  3d-aware image synthesis.
\newblock In {\em IEEE CVPR}, pages 5799--5809, 2021.

\bibitem{chen2021mvsnerf}
Anpei Chen, Zexiang Xu, Fuqiang Zhao, Xiaoshuai Zhang, Fanbo Xiang, Jingyi Yu,
  and Hao Su.
\newblock Mvsnerf: Fast generalizable radiance field reconstruction from
  multi-view stereo.
\newblock {\em arXiv preprint arXiv:2103.15595}, 2021.

\bibitem{cosmo2020limp}
Luca Cosmo, Antonio Norelli, Oshri Halimi, Ron Kimmel, and Emanuele Rodol{\`a}.
\newblock Limp: Learning latent shape representations with metric preservation
  priors.
\newblock In {\em ECCV}, pages 19--35. Springer, 2020.

\bibitem{unconstrained-scene-generation}
Terrance DeVries, Miguel~Angel Bautista, Nitish Srivastava, Graham~W. Taylor,
  and Joshua~M. Susskind.
\newblock Unconstrained scene generation with locally conditioned radiance
  fields.
\newblock 2021.

\bibitem{dinh2014nice}
Laurent Dinh, David Krueger, and Yoshua Bengio.
\newblock Nice: Non-linear independent components estimation.
\newblock {\em arXiv preprint arXiv:1410.8516}, 2014.

\bibitem{dinh2016density}
Laurent Dinh, Jascha Sohl-Dickstein, and Samy Bengio.
\newblock Density estimation using real nvp.
\newblock {\em arXiv preprint arXiv:1605.08803}, 2016.

\bibitem{gadelha2020learning}
Matheus Gadelha, Giorgio Gori, Duygu Ceylan, Radomir Mech, Nathan Carr, Tamy
  Boubekeur, Rui Wang, and Subhransu Maji.
\newblock Learning generative models of shape handles.
\newblock In {\em IEEE CVPR}, pages 402--411, 2020.

\bibitem{garbin2021fastnerf}
Stephan~J Garbin, Marek Kowalski, Matthew Johnson, Jamie Shotton, and Julien
  Valentin.
\newblock Fastnerf: High-fidelity neural rendering at 200fps.
\newblock {\em arXiv preprint arXiv:2103.10380}, 2021.

\bibitem{grover2018flow}
Aditya Grover, Manik Dhar, and Stefano Ermon.
\newblock Flow-gan: Combining maximum likelihood and adversarial learning in
  generative models.
\newblock In {\em AAAI}, 2018.

\bibitem{gu2021stylenerf}
Jiatao Gu, Lingjie Liu, Peng Wang, and Christian Theobalt.
\newblock Stylenerf: A style-based 3d-aware generator for high-resolution image
  synthesis.
\newblock {\em arXiv preprint arXiv:2110.08985}, 2021.

\bibitem{fast-and-explicit-neural-view-synthesis}
Pengsheng Guo, Miguel~Angel Bautista, Alex Colburn, Liang Yang, Daniel
  Ulbricht, Joshua~M. Susskind, and Qi Shan.
\newblock Fast and explicit neural view synthesis, 2021.

\bibitem{guo2020materialgan}
Yu Guo, Cameron Smith, Milo{\v{s}} Ha{\v{s}}an, Kalyan Sunkavalli, and Shuang
  Zhao.
\newblock Materialgan: reflectance capture using a generative svbrdf model.
\newblock {\em arXiv preprint arXiv:2010.00114}, 2020.

\bibitem{harkonen2020ganspace}
Erik H{\"a}rk{\"o}nen, Aaron Hertzmann, Jaakko Lehtinen, and Sylvain Paris.
\newblock Ganspace: Discovering interpretable gan controls.
\newblock {\em arXiv preprint arXiv:2004.02546}, 2020.

\bibitem{hedman2021baking}
Peter Hedman, Pratul~P Srinivasan, Ben Mildenhall, Jonathan~T Barron, and Paul
  Debevec.
\newblock Baking neural radiance fields for real-time view synthesis.
\newblock {\em arXiv preprint arXiv:2103.14645}, 2021.

\bibitem{henzler2020neuraltexture}
Philipp Henzler, Niloy~J Mitra, , and Tobias Ritschel.
\newblock Learning a neural 3d texture space from 2d exemplars.
\newblock In {\em CVPR}, June 2019.

\bibitem{henzler2019escaping}
Philipp Henzler, Niloy~J Mitra, and Tobias Ritschel.
\newblock Escaping plato's cave: 3d shape from adversarial rendering.
\newblock In {\em ICCV}, pages 9984--9993, 2019.

\bibitem{higgins2016beta}
Irina Higgins, Loic Matthey, Arka Pal, Christopher Burgess, Xavier Glorot,
  Matthew Botvinick, Shakir Mohamed, and Alexander Lerchner.
\newblock beta-vae: Learning basic visual concepts with a constrained
  variational framework.
\newblock 2016.

\bibitem{hinz2021improved}
Tobias Hinz, Matthew Fisher, Oliver Wang, and Stefan Wermter.
\newblock Improved techniques for training single-image gans.
\newblock In {\em Proceedings of the IEEE/CVF Winter Conference on Applications
  of Computer Vision}, pages 1300--1309, 2021.

\bibitem{isola2017image}
Phillip Isola, Jun-Yan Zhu, Tinghui Zhou, and Alexei~A Efros.
\newblock Image-to-image translation with conditional adversarial networks.
\newblock In {\em IEEE CVPR}, pages 1125--1134, 2017.

\bibitem{jolicoeur2020adversarial}
Alexia Jolicoeur-Martineau, R{\'e}mi Pich{\'e}-Taillefer, R{\'e}mi Tachet~des
  Combes, and Ioannis Mitliagkas.
\newblock Adversarial score matching and improved sampling for image
  generation.
\newblock {\em arXiv preprint arXiv:2009.05475}, 2020.

\bibitem{kar2019meta}
Amlan Kar, Aayush Prakash, Ming-Yu Liu, Eric Cameracci, Justin Yuan, Matt
  Rusiniak, David Acuna, Antonio Torralba, and Sanja Fidler.
\newblock Meta-sim: Learning to generate synthetic datasets.
\newblock In {\em ICCV}, pages 4551--4560, 2019.

\bibitem{Karnewar2022ReLUFields}
Animesh Karnewar, Tobias Ritschel, Oliver Wang, and Niloy~J. Mitra.
\newblock {ReLU} fields: The little non-linearity that could.
\newblock In {\em Proc. of {SIGGRAPH}}, volume~41, pages 13:1--13:8, 2022.

\bibitem{karnewar2020msg}
Animesh Karnewar and Oliver Wang.
\newblock Msg-gan: Multi-scale gradients for generative adversarial networks.
\newblock In {\em IEEE CVPR}, pages 7799--7808, 2020.

\bibitem{karras2017progressive}
Tero Karras, Timo Aila, Samuli Laine, and Jaakko Lehtinen.
\newblock Progressive growing of gans for improved quality, stability, and
  variation.
\newblock {\em arXiv preprint arXiv:1710.10196}, 2017.

\bibitem{karras2020training}
Tero Karras, Miika Aittala, Janne Hellsten, Samuli Laine, Jaakko Lehtinen, and
  Timo Aila.
\newblock Training generative adversarial networks with limited data.
\newblock {\em arXiv preprint arXiv:2006.06676}, 2020.

\bibitem{karras2021alias}
Tero Karras, Miika Aittala, Samuli Laine, Erik H{\"a}rk{\"o}nen, Janne
  Hellsten, Jaakko Lehtinen, and Timo Aila.
\newblock Alias-free generative adversarial networks.
\newblock {\em arXiv preprint arXiv:2106.12423}, 2021.

\bibitem{karras2019style}
Tero Karras, Samuli Laine, and Timo Aila.
\newblock A style-based generator architecture for generative adversarial
  networks.
\newblock In {\em IEEE CVPR}, pages 4401--4410, 2019.

\bibitem{karras2020analyzing}
Tero Karras, Samuli Laine, Miika Aittala, Janne Hellsten, Jaakko Lehtinen, and
  Timo Aila.
\newblock Analyzing and improving the image quality of stylegan.
\newblock In {\em IEEE CVPR}, pages 8110--8119, 2020.

\bibitem{kim2021drivegan}
Seung~Wook Kim, Jonah Philion, Antonio Torralba, and Sanja Fidler.
\newblock Drivegan: Towards a controllable high-quality neural simulation.
\newblock In {\em IEEE CVPR}, pages 5820--5829, 2021.

\bibitem{kingma2018glow}
Diederik~P Kingma and Prafulla Dhariwal.
\newblock Glow: Generative flow with invertible 1x1 convolutions.
\newblock {\em arXiv preprint arXiv:1807.03039}, 2018.

\bibitem{kingma2013auto}
Diederik~P Kingma and Max Welling.
\newblock Auto-encoding variational bayes.
\newblock {\em arXiv preprint arXiv:1312.6114}, 2013.

\bibitem{pmlr-v48-larsen16}
Anders Boesen~Lindbo Larsen, Søren~Kaae Sønderby, Hugo Larochelle, and Ole
  Winther.
\newblock Autoencoding beyond pixels using a learned similarity metric.
\newblock volume~48, pages 1558--1566, New York, New York, USA, 20--22 Jun
  2016.

\bibitem{max1995optical}
Nelson Max.
\newblock Optical models for direct volume rendering.
\newblock {\em IEEE Transactions on Visualization and Computer Graphics},
  1(2):99--108, 1995.

\bibitem{mildenhall2019local}
Ben Mildenhall, Pratul~P Srinivasan, Rodrigo Ortiz-Cayon, Nima~Khademi
  Kalantari, Ravi Ramamoorthi, Ren Ng, and Abhishek Kar.
\newblock Local light field fusion: Practical view synthesis with prescriptive
  sampling guidelines.
\newblock {\em ACM Trans. Graph.}, 38(4):1--14, 2019.

\bibitem{mildenhall2020nerf}
Ben Mildenhall, Pratul~P Srinivasan, Matthew Tancik, Jonathan~T Barron, Ravi
  Ramamoorthi, and Ren Ng.
\newblock Nerf: Representing scenes as neural radiance fields for view
  synthesis.
\newblock In {\em ECCV}, pages 405--421, 2020.

\bibitem{nash2020polygen}
Charlie Nash, Yaroslav Ganin, SM~Ali Eslami, and Peter Battaglia.
\newblock Polygen: An autoregressive generative model of 3d meshes.
\newblock In {\em International Conference on Machine Learning}, pages
  7220--7229. PMLR, 2020.

\bibitem{nguyen2020rgbd}
Phong Nguyen, Animesh Karnewar, Lam Huynh, Esa Rahtu, Jiri Matas, and Janne
  Heikkila.
\newblock Rgbd-net: Predicting color and depth images for novel views
  synthesis.
\newblock {\em arXiv preprint arXiv:2011.14398}, 2020.

\bibitem{nguyen2019hologan}
Thu Nguyen-Phuoc, Chuan Li, Lucas Theis, Christian Richardt, and Yong-Liang
  Yang.
\newblock Hologan: Unsupervised learning of 3d representations from natural
  images.
\newblock In {\em ICCV}, pages 7588--7597, 2019.

\bibitem{nguyen2020blockgan}
Thu Nguyen-Phuoc, Christian Richardt, Long Mai, Yong-Liang Yang, and Niloy
  Mitra.
\newblock Blockgan: Learning 3d object-aware scene representations from
  unlabelled images.
\newblock {\em arXiv preprint arXiv:2002.08988}, 2020.

\bibitem{niemeyer2021giraffe}
Michael Niemeyer and Andreas Geiger.
\newblock Giraffe: Representing scenes as compositional generative neural
  feature fields.
\newblock In {\em IEEE CVPR}, pages 11453--11464, 2021.

\bibitem{Ost_2021_CVPR}
Julian Ost, Fahim Mannan, Nils Thuerey, Julian Knodt, and Felix Heide.
\newblock Neural scene graphs for dynamic scenes.
\newblock In {\em IEEE CVPR}, pages 2856--2865, June 2021.

\bibitem{pidhorskyi2020adversarial}
Stanislav Pidhorskyi, Donald~A Adjeroh, and Gianfranco Doretto.
\newblock Adversarial latent autoencoders.
\newblock In {\em IEEE CVPR}, pages 14104--14113, 2020.

\bibitem{reiser2021kilonerf}
Christian Reiser, Songyou Peng, Yiyi Liao, and Andreas Geiger.
\newblock Kilonerf: Speeding up neural radiance fields with thousands of tiny
  mlps.
\newblock {\em arXiv preprint arXiv:2103.13744}, 2021.

\bibitem{schonberger2016structure}
Johannes~L Schonberger and Jan-Michael Frahm.
\newblock Structure-from-motion revisited.
\newblock In {\em IEEE CVPR}, pages 4104--4113, 2016.

\bibitem{schwarz2020graf}
Katja Schwarz, Yiyi Liao, Michael Niemeyer, and Andreas Geiger.
\newblock Graf: Generative radiance fields for 3d-aware image synthesis.
\newblock {\em arXiv preprint arXiv:2007.02442}, 2020.

\bibitem{shaham2019singan}
Tamar~Rott Shaham, Tali Dekel, and Tomer Michaeli.
\newblock Singan: Learning a generative model from a single natural image.
\newblock In {\em ICCV}, pages 4570--4580, 2019.

\bibitem{shen2020interfacegan}
Yujun Shen, Ceyuan Yang, Xiaoou Tang, and Bolei Zhou.
\newblock Interfacegan: Interpreting the disentangled face representation
  learned by gans.
\newblock {\em IEEE Trans. Pattern Anal. Mach. Intell.}, 2020.

\bibitem{song2021maximum}
Yang Song, Conor Durkan, Iain Murray, and Stefano Ermon.
\newblock Maximum likelihood training of score-based diffusion models.
\newblock {\em arXiv preprint arXiv:2101.09258}, 2021.

\bibitem{song2020improved}
Yang Song and Stefano Ermon.
\newblock Improved techniques for training score-based generative models.
\newblock {\em arXiv preprint arXiv:2006.09011}, 2020.

\bibitem{tewari2020state}
Ayush Tewari, Ohad Fried, Justus Thies, Vincent Sitzmann, Stephen Lombardi,
  Kalyan Sunkavalli, Ricardo Martin-Brualla, Tomas Simon, Jason Saragih,
  Matthias Nie{\ss}ner, et~al.
\newblock State of the art on neural rendering.
\newblock In {\em Comput. Graph. Forum}, volume~39, pages 701--727, 2020.

\bibitem{vahdat2020nvae}
Arash Vahdat and Jan Kautz.
\newblock Nvae: A deep hierarchical variational autoencoder.
\newblock {\em arXiv preprint arXiv:2007.03898}, 2020.

\bibitem{wang2018high}
Ting-Chun Wang, Ming-Yu Liu, Jun-Yan Zhu, Andrew Tao, Jan Kautz, and Bryan
  Catanzaro.
\newblock High-resolution image synthesis and semantic manipulation with
  conditional gans.
\newblock In {\em IEEE CVPR}, pages 8798--8807, 2018.

\bibitem{wu2016learning}
Jiajun Wu, Chengkai Zhang, Tianfan Xue, William~T Freeman, and Joshua~B
  Tenenbaum.
\newblock Learning a probabilistic latent space of object shapes via 3d
  generative-adversarial modeling.
\newblock In {\em Adv. Neural Inform. Process. Syst.}, pages 82--90, 2016.

\bibitem{yu2021plenoctrees}
Alex Yu, Ruilong Li, Matthew Tancik, Hao Li, Ren Ng, and Angjoo Kanazawa.
\newblock Plenoctrees for real-time rendering of neural radiance fields.
\newblock {\em arXiv preprint arXiv:2103.14024}, 2021.

\bibitem{yu2021pixelnerf}
Alex Yu, Vickie Ye, Matthew Tancik, and Angjoo Kanazawa.
\newblock pixelnerf: Neural radiance fields from one or few images.
\newblock In {\em IEEE CVPR}, pages 4578--4587, 2021.

\bibitem{zhang2020nerf++}
Kai Zhang, Gernot Riegler, Noah Snavely, and Vladlen Koltun.
\newblock Nerf++: Analyzing and improving neural radiance fields.
\newblock {\em arXiv preprint arXiv:2010.07492}, 2020.

\bibitem{zhang2020image}
Yuxuan Zhang, Wenzheng Chen, Huan Ling, Jun Gao, Yinan Zhang, Antonio Torralba,
  and Sanja Fidler.
\newblock Image gans meet differentiable rendering for inverse graphics and
  interpretable 3d neural rendering.
\newblock {\em arXiv preprint arXiv:2010.09125}, 2020.

\bibitem{zhou2021cips}
Peng Zhou, Lingxi Xie, Bingbing Ni, and Qi Tian.
\newblock Cips-3d: A 3d-aware generator of gans based on
  conditionally-independent pixel synthesis.
\newblock {\em arXiv preprint arXiv:2110.09788}, 2021.

\bibitem{zhou2018stereo}
Tinghui Zhou, Richard Tucker, John Flynn, Graham Fyffe, and Noah Snavely.
\newblock Stereo magnification: Learning view synthesis using multiplane
  images.
\newblock In {\em ACM Trans. Graph.}, 2018.

\end{thebibliography}
}
\end{document}